\documentclass{article}
\usepackage{cleveref}
\usepackage{icml2005}
\usepackage{ella}
\usepackage{mlapa}
\usepackage{graphicx}
\usepackage{amsmath}
\usepackage{amssymb}
\usepackage[english,polish]{babel}

\begin{document}

\twocolumn[ \icmltitle{Domain based classification}

\icmlauthor{Robert P.W. Duin}{r.p.w.duin@ewi.tudelft.nl}
\icmladdress{ICT group, Faculty of Electr. Eng., Mathematics and Computer Science\\
Delft University of Technology, The Netherlands}
\icmlauthor{Elzbieta Pekalska}{e.pekalska@ewi.tudelft.nl}
\icmladdress{ICT group, Faculty of Electr. Eng., Mathematics and Computer Science\\
Delft University of Technology, The Netherlands}

\vskip 0.3in]

\begin{abstract}
The majority of traditional classification rules minimizing the expected
probability of error (0-1 loss) are inappropriate if the class probability
distributions are ill-defined or impossible to estimate. We argue that in such
cases class domains should be used instead of class distributions or densities
to construct a reliable decision function. Proposals are presented for some
evaluation criteria and classifier learning schemes, illustrated by an example.
\end{abstract}

\section{Introduction} \label{introduction}
Probabilistic framework is often employed to solve learning problems. One
conveniently assumes that real-world objects or phenomena are represented as
(or, in fact, reduced to) vectors $\bx$ in a suitable vector space $\XX$. The
learning task relies on finding an unknown functional dependency between $\bx$
and some outputs $y \inn \YY$. Vectors $\bx$ are assumed to be iid, \ie drawn
independently from a fixed, but unknown probability distribution $p(\bx)$. The
function $f$ is given as a fixed conditional density $p(y|\bx)$, which is also
unknown. To search for the ideal function $f^*$, a general space of hypothesis
functions $\FF = \{f\!: \XX \!\to\! \YY\}$ is considered. $f^*$ is considered
optimal according to some loss function $L\!: \XX \!\times\! \YY \!\to\!
[0,M]$, $M \g 0$, measuring the discrepancy between the true and estimated
values. The learning problem is then formulated as minimizing the true error
$\EE(f) \eq \int_{\XX \times \YY} L(y,f(\bx)) p(\bx,y) d\bx\, dy$, given a
finite iid sample, \ie the training set $\{(\bx_i,y_i)\}$, $i\eq 1,2,\ldots,n$.
As the joint probability $p(\bx,y) \eq p(\bx) p(y|\bx)$ is unknown, one,
therefore, minimizes the empirical error $\EE_{emp}(f) \eq \frac{1}{n}
\sum_{i=1}^n L(y_i,f(\bx_i))$. Additionally, a trade-off between the function
complexity and the fit to the data has to be kept, as a small empirical error
does not yet guarantee a small true error. This is achieved by adding a
suitable penalty or regularization function as proposed in the structural risk
minimization or regularization principles.

Although these principles are mathematically well-founded, they rely on very
strong, though general, assumptions. They impose a fixed (stationary)
distribution from which vectors, representing objects, are drawn. Moreover, the
training set is believed to be representative for the task. Usually, it is a
random subset of some large target set, such as a set of all objects in an
application. Such assumptions are often violated in practice, not only due to
differences in measurements caused by variability between sensors or a
difference in calibration of measuring devices, but, more importantly, due to
the lack of information on class distributions or impossibility of gathering a
representative sample. Some examples are:
\begin{itemize}
\item
In the application of face detection, the distribution of non-faces cannot be
determined, as it may be unknown for which type of images and in which
environments such a detector is going to be used.
\item
In machine diagnostics and industrial inspection some of the classes have to be
artificially induced in order to obtain sufficient examples for training.
Whether they reflect the true distribution may be unknown.
\item
In geological exploration for mining purposes, a large set of examples may be
easily obtained in one area on earth, but its distribution might be entirely
different than in another area, whose sample will not be provided due to the
related costs.
\end{itemize}
In human learning, a random sampling of the distribution of the target set does
not seem to be a plausible approach, as it is usually not very helpful to
encounter multiple copies of the same object among training examples. For
instance, in studying the difference between the paintings of Rembrandt and
Velasquez it makes no sense to consider copies of the same painting. Even very
similar ones may be superfluous, in spite of the fact that they may represent
some mode in the distribution of paintings. On the contrary, it may be better
to emphasize the tails or the borders of the distribution, especially in the
situations, where the classes seem to be hard to distinguish.

Although the probabilistic framework is applied to many learning problems,
there are many practical situations, where alternative paradigms are necessary
due to the nature of ill-sampled data or ill-defined distributions. Which may
be an appropriate model for the relation between a training set of examples and
the target set of objects to be classified\footnote{In classification problems,
$y_i$ is a class label and $L$ is the 0-1 loss, $L(y_i,f(\bx_i)) \eq
\II(y_i\!\neq \!f(\bx_i))$, where $\II$ is the indicator function. Classifiers
minimize the expected classification error (0-1 loss). } if we cannot or do not
want to assume that the distribution of the training set is an approximation of
the distribution of the target set?  This paper focusses on this aspect. Our
basic assumption is that the training sample is representative for the
\emph{domain} of the target set (all examples in the given application) instead
of being drawn from a \emph{fixed probability distribution}.

Consider a representation space, called also input space, $\XX$, endowed with
some metric $d$. This is the space, in which objects are represented as vectors
and the learning takes place. A \emph{domain} is a bounded set $\A$ in $\XX$,
\ie $\exists r\g 0 ~\forall x,z\inn \A  ~d(x,z) \lee r$. (We do not assume that
the domain is totally bounded.) This is not new as one usually expects that
classes are represented by a set of vectors in (possibly convex and) bounded
subsets of some space. Here, we will focus on vector space representations
constructed by features, dissimilarities or kernels. As the class domains are
bounded in this representation, for each class $\omega_j$, there exists some
indicator function $G_j(\bx)$ of the object\footnote{By an object we mean its
representation $\bx$ in the considered vector space.} $\bx$ such that
$G_j(\bx)\eq 1$ if $\bx$ is accepted as a member of $\omega_j$ and $G_j(\bx)\eq
0$, otherwise. Given a training set of labeled examples $\{(\bx_i,y_i)\}$,
$G_j(\bx_i)\eq 1$ if $\bx_i$ belongs to the class $\omega_j$. We will assume
that each object belongs to a single class, however, identical objects with
different labels are permitted. This allows classes to overlap.

Given the above model, several questions arise. How to design learning
procedures and how to evaluate them? Can classifiers output confidences? How to
judge whether a given training set is representative for the domain? Are any
further assumptions needed or advantageous? Can cluster analysis or feature
selection be applied? The goal of this paper is to raise interest in domain
learning. As the first step, we introduce the problem, discuss a few issues and
propose some approaches.


\section{Performance criteria} \label{performance}
Suppose a classifier $f(\bx)$ is designed that assigns objects to one of the
given classes. A labeled evaluation set or a test set $S$ is usually used to
estimate the performance of $f(\bx)$ by counting the number of incorrect
assignments. This, however, demands that the set $S$ is representative for the
distribution of the target set, which conflicts with our assumption.

For a set of objects to be representative for the class domains it may be
assumed that the objects are well spread over these domains. For the test set
$S$, it means that there is no object $\bx$ in any of the classes that has a
large distance $d(\bx,\bx^s)$ to its nearest objects $\bx^s \inn S$. Therefore,
for a domain representative test set $S$ holds that
\begin{equation}
d_{max} = \max_{\bx} ~\min_{\bx^s \in S} ~d(\bx,\bx^s)
\end{equation}
is small $\forall \bx$. The usefulness of this approach relies on the fact that
the distances as given in the input space are meaningful for the application.
Consequently, for a well-performing classifier, none of the erroneously
classified objects is far away (at the wrong side) from the decision boundary.
If the classes are separable, the test objects should also be as far away from
the decision boundary as possible. Therefore, our proposal is to follow the
worst-case scenario and to judge a classifier by the object that is the most
misleading. This will be judged by its distance to the decision boundary.

Consider a two-class problem with the labels $y \inn \{-\!1,+\!1\}$, where
$y(\bx)$ denotes the true label of $\bx$. (This notation is the consequence of
our assumption that different objects with different labels may be represented
in the same point $\bx$). Let $f(\bx)$ yield the signed distance of $\bx$ to
the decision boundary induced by the classifier. Note that the unsigned
distance of $\bx$ to the decision boundary is related to the functional form
of~$f$. Then
\begin{equation}
\eta(S|g) = \min_{\bx^s \in S} ~y(\bx^s)f(\bx^s)
\label{eq:criterion}
\end{equation}
is the signed distance to the decision boundary of the 'worst' classified
object from the test set $S$. Having introduced this, a classifier $f_1(\bx)$
is judged to be better than a classifier $f_2(\bx)$ if $\eta(S|f_1) \g
\eta(S|f_2)$. The main argument supporting this criterion follows from the fact
that if the vector space representation and the distance measure are
appropriate for the learning problem, then for small values of $\eta(S|f)$, the
test set $S$ contains objects that are similar to the objects in a wrong class.
As the data and the learning procedure are not based on probabilities, it is
difficult to make a statement about the probability of errors instead of the
seriousness of their contributions.

As a consequence, outliers should be avoided, since they cannot be detected by
statistical means. Still, objects that have large distances to all other
objects (in comparison to their nearest neighbor distances) indicate that the
domain is not well sampled. If the sampling is proper, all objects have to be
considered as equally important, as they are examples of valid representations.
Copies of the same object do not influence the learning procedures and may,
therefore, be removed.

If classes overlap such that the overlapping domain can be estimated and a
class of possible density functions is provided, then it might be possible to
determine generalization bounds for the classification error or to estimate the
expected error over the class of density functions. Both tasks are, however,
not straightforward, neither estimating the domain of the class overlap, nor
defining an appropriate class of density functions. As we only sketch the
problem, we will restrict ourselves to classifiers that maximize
criterion~\eqref{eq:criterion}.

\section{Classifier proposals}\label{classifiers}
A number of possible domain based decision functions will be introduced in this
section. We will start by presenting the domain versions of some well-known
probabilistic classifiers. It should be emphasized once again that in the
probabilistic framework, any averaging over objects or their functional
dependencies relies on their distribution. So, averaging cannot be used in
domain based learning procedures. It has to be replaced by appropriate
operators such as minimum, maximum or domain center.

Consider a vector space $\RR^m$, in which objects $\bx$ are represented \eg by
features. Let $X \eq \{\bx_1,\bx_2, \ldots,\bx_n\}$ be a training set with the
labels $Y \eq \{y_1,y_2, \ldots,y_n\}$. Assume $k$ classes
$\omega_1,\ldots,\omega_k$. If $k\eq 2$, then $y_i \inn\{-\!1,+\!1\}$ are
assumed. Let $X_j$ be a subset of $X$ containing all members of $\omega_j$.
Then, $X \eq \bigcup_{j} X_j$.

\subsection{Discriminants}\label{subs:discriminants}
Consider a two-class problem. If classes are separable by a polynomial or when
a kernel transformation is applied, a discriminant function can be found by
solving a set of linear inequalities over the training set $X$, \eg
\begin{equation}
 y_i(\bmg{\alpha}^T K(X,x_i)+\alpha_0) > 0, \quad \forall \bx_i \inn X,
\label{eq:lin_inequal}
\end{equation}
$K(X,\bx_i)$ is the column vector of all kernel values $K(\bx,\bx_i)$, $\forall
\bx \inn X$. The resulting weights $\bmg{\alpha} \inn \RR^n$ define the
classifier $f$ in the following way:
\begin{equation}
\label{eq:discriminant}
\begin{split}
\text{Assign $\bx$ to~} \omega_1, \text{if~} f(\bx) &= \bmg{\alpha}^T K(X,\bx)+\alpha_0 \ge 0,\\
\text{Assign $\bx$ to~} \omega_2, \text{if~} f(\bx) &= \bmg{\alpha}^T
K(X,\bx)+\alpha_0 < 0.
\end{split}
\end{equation}
This decision function finds a solution if the classes are separable in the
Hilbert space induced by the kernel $K$ and fails if they are not. Since no
model used to optimize the decision boundary, this decision function is
independent of the use of domains or densities.

In the traditional probabilistic approach to pattern recognition, the nearest
mean classifier (NMC) and Fisher's linear discriminant (FLD) are two frequently
used classifiers. Given class means estimated over the training set, the NMC
assigns each object to the class of its nearest mean. In a domain approach,
class means should be replaced by the class centers. These are vectors
$\bmg{\mu}_j$ in the vector space $\RR^m$ that yield the minimum distance to
the most remote object in $X_j$:
\begin{equation}
\hat{\bmg{\mu}}_j = \arg\min_{\bx^* \inn \RR^m}  ~\max_{\bx \in X_j}
\|\bx-\bx^*\| \label{eq:def_center}
\end{equation}
Class centers may be found by a procedure like the Support Vector Data
Description \cite{TaxDui1999a,Tax2001}, in agreement to criterion
\eqref{eq:def_center}. Such a center is determined by  $m\!+\!1$ training
objects at most, and usually much less. An approximation can be also based on a
feature-by-feature computation. Additionally, for sufficiently large data, a
single training object may be a sufficiently good approximation of the center:
\begin{equation}
\hat{\bmg{\mu}}_j = \arg\min_{\bx^* \inn X} ~\max_{\bx \in X_j} \|\bx-\bx^*\|.
\label{eq:nearest_center_estimate}
\end{equation}
This can be determined fast from the pairwise distance matrix
computed between the training examples \cite{Hochbaum85}. Given
the class centers, the Nearest Center Classifier (NCC) is now
defined as:
\begin{equation}
\text{Assign $\bx$ to~} \omega_i, \text{if~}
i=\arg\min_j~\|x-\hat{\bm{\mu}_j}\|. \label{eq:ncc}
\end{equation}
This classifier is optimal (it maximizes criterion \ref{eq:criterion}) if the
class domains are hyperspheres with identical radii.

A traditional criterion for judging the goodness of a single feature
is the Fisher Criterion:
\begin{equation}
J_F = \frac{(\mu_1-\mu_2)^2}{{\sigma_1}^2+{\sigma_2}^2}
\label{eq:Fisher_crit}
\end{equation}
in which $\mu_j$ and $\sigma_j^2$ are the class means and variances,
respectively, as computed for the single feature. A domain based version is
defined by substituting the mean with the class center and the variance with
the squared class range. For the $k$-th feature, ${\sigma_j}^2$ can be then
estimated as:
\begin{equation}
\hat{\sigma}_j^2 = (\max_i(\bx_{ik})-\min_i (\bx_{ik}))^2
\label{eq:class_width}
\end{equation}
Herewith, a Fisher Linear Domain Discriminant (FLDD) can be defined by a weight
vector in the feature space for which the domain version of
\eqref{eq:Fisher_crit} is maximum. We expect that this direction will be
determined by the minimum-volume ellipsoid enclosing $X^{c}$, the pooled data
shifted by the class centers $X_j^{c} \eq \{\bx \!-\! \bmg{\mu}_j\!: \bx \inn
X_j\}$. It is defined by the positive semi-definite matrix $G$, such that
$\bx^T G \bx < 1, \forall \bx \inn X^c$. Consequently, one has:
\begin{equation}
\begin{array}{l}
\text{Assign $\bx$ to~} \omega_1,  \\[0.5mm]
\text{if~} (\bx-\bmg{\mu}_2)^T G^{-1} (\bx-\bmg{\mu}_2) \ge (\bx-\bmg{\mu}_1)^T G^{-1} (\bx-\bmg{\mu}_1),\\[1mm]
\text{Assign $\bx$ to~} \omega_2,  \\[0.5mm]
\text{if~} (\bx-\bmg{\mu}_2)^T G^{-1} (\bx-\bmg{\mu}_2) < (\bx-\bmg{\mu}_1)^T
G^{-1} (\bx-\bmg{\mu}_1).
\end{array}
\end{equation}
The FLDD can then be written as:
\begin{equation}
f(\bx) = (\bmg{\mu}_2 - \bmg{\mu}_1)^T G^{-1} \bx.
\label{eq:FLDD}
\end{equation}
This classifier is optimal according to criterion \eqref{eq:criterion} if the
two classes are described by the identical ellipsoids except for the position
of their centers. The estimation of $G$ in the problem of finding the minimum
volume ellipsoid enclosing the data $X$ is a convex optimization problem which
is only tractable in special cases \cite{BerBoy1996,BoyBer2004}. An
approximation is possible when the joint covariance matrix is used for
pre-whitening the data (which, however, conflicts with the concept of a domain
classifier) and then deriving a hypersphere instead of an ellipsoid.

As a third possibility in this section we will mention the binary decision tree
classifier based on the purity criterion \cite{Breiman84}, capturing aspects of
partitioning of examples relevant to good classification. In each node of the
tree, the feature and a threshold are determined to distinguish the largest
pure part (\ie a range belonging to just one of the classes) of the training
set.  Other more advanced ways of finding a domain based learner will be
discussed below.

\subsection{Model based, parametric decision functions}
Two of the methods described in the previous section aim at finding
discriminants by some separability criterion such as the difference in class
centers or the Fisher distance. They appear to be optimal for identically
shaped class domains, hyperspheres and, respectively, ellipsoids. Here, instead
of considering a functional form of a classifier, we will start from some class
domain models and then determine the classifier.

Class domains are defined by their boundaries. If during a training process
some objects are placed outside the domain, the boundaries have to be adjusted.
This is permitted only if the nearest objects inside the domain are close to
the boundaries or their parts (if distinguishable). 'Unreasonably far away'
objects should not play a role in positioning of the domain boundaries. They
have to be determined with respect to the demand that objects should sample the
domain well. So, the distance from the domain boundary to the nearest objects
should be comparable to the nearest neighbor distances between the objects. In
fact, this is the basic learning problem \cite{Valiant84}. A significant
difference to many later studies \cite{KulkarniZ93}, however, is that in domain
learning probabilities or densities cannot be used.

Formally, the problem may be stated as follows. Let $D_j(\bx,\bmg{\theta})\eq
0$ be some parametric domain description (with the parameters $\bmg{\theta}$
for the class $\omega_j$ and let $X_j$ be a set of examples from $\omega_j$.
Then, $\bmg{\theta}$ should be chosen such that the maximum distance from the
domain boundary to its nearest neighbor in the training set is minimized under
the condition that all training objects are inside the domain at some suitable
distance $\delta$ to the border:
\begin{equation}
\label{eq:model_crit}
\begin{array}{ll}
\min_{\bmg{\theta}} & \max_{\bx^*} ~\min_{\bx \in X_j} ~\|\bx^* -\bx\|,\\[1mm]
\text{s.t.}         & (a) ~ D(\bx^*,\bmg{\theta})\eq 0,  \\[1mm]
\text{s.t.}         & (b) ~D_j(\bx,\bmg{\theta}) < 0, ~~\forall \bx \inn X_j \\[1mm]
                    & (c) ~\|\bx^* - \bx\| > \delta, ~~\forall \bx \inn X_j
\end{array}
\end{equation}
This is a nonlinear optimization. As indicated above, such problems are
intractable already for simple domains like arbitrary ellipsoids
\cite{BoyBer2004}. The challenge, therefore, is to find approximate and
feasible solutions. Examples can be found in the area of one-class classifiers
\cite{Tax2001,Scholkopf01}. A very problematic issue, however, is the
constraint (c) in \eqref{eq:model_crit} indicating that the domain border
should fit loosely, but in a restricted way around the training examples in the
feature space. The difficulty arises as $\|\bx^* - \bx\| \g \delta$ is a
non-convex constraint, hence the entire formulation is
non-convex\footnote{Convex optimization deals with a well-behaved set of
problems that have advantageous theoretical properties such as the duality
theory and for which efficient algorithms exist. This is not true for
non-convex problems.}. In domain learning, new algorithms have to be designed
to solve the formulated problems.

Once class domains have been found, the problem of a proper class assignment
arises if objects get multiple memberships or if they are rejected by all
classes. If a unique decision is demanded in such cases, a discriminant has to
be determined, as discussed in section \ref{subs:discriminants}. Alternatively,
during classification, the distances to all domain boundaries have to be found
and the smallest, in the case of reject, or the largest, in the case of
multiple acceptance,  has to be used for the final decision. Again, the
criterion \eqref{eq:criterion} is used.

\subsection{Model based, non-parametric decision functions}
Instead of estimating the parameters of some postulated model, such a model
might be also directly constructed from the training set, in analogy to the
kernel density (Parzen) estimators \cite{Parzen62} in statistical learning. For
a domain description, the sum of kernel functions, however, may be replaced by
a maximum, or, equivalently, by the union of the kernel domains. In order to
restrict the class domains, the kernel domain should be bounded. Let
$\Phi(\bx,\bx_i,h)$ define the domain for a kernel associated with $\bx_i$, \eg
all points within a hypersphere with the radius $h$, then the domain estimate
for the class $\omega_j$ is:
\begin{equation}
  D_j(\bx,h) = \bigcup_{\bx_i \in X_j} \{\Phi(\bx,\bx_i,h)\}.
\end{equation}
The value of the kernel width $h$ can be estimated by the leave-one-out
procedure. $h$ is found as the smallest value for which all training objects
belong to the domain which is estimated by all training objects except the one
to be classified. This width is equal to the largest nearest neighbor distance
found in the training set:
\begin{equation}\label{eq:kernel_width}
    \hat{h} = \max_i ~\min_{l\neq i}~\|\bx_i-\bx_l\|.
\end{equation}
Also in this case it is not straightforward how the distance to the domain
boundary should be computed.

\subsection{Neural networks}
The iterative way neural networks are trained make them suitable for domain
learning. Traditionally, the weights of a neural network are chosen to minimize
the mean square error over the training set \cite{Bishop95}:
\begin{equation}\label{eq:neural_net1}
  \hat{\bmg{\theta}} = \arg\min_{\bmg{\theta}}
 ~\frac{1}{n} \sum_{\bx \in X}(\textrm{net}(\bx,\bmg{\theta})-t(\bx))^2,
\end{equation}
where $\textrm{net}(\bx,\bmg{\theta})$ is the network output for $\bx$ and
$t(\bx)$ is the target, which is $y(\bx)$ here. As the network function is
nonlinear, training is done in small steps following a gradient descent
approach. The summation over the training examples, however, conflicts with the
domain learning idea. If it is replaced by the maximum operator, the network
will be updated such that the 'worst' object, \ie the object closest to the
domain of the other class, makes as smallest error as possible (it is as close
as possible to the decision border):
\begin{equation}\label{eq:neural_net2}
  \hat{\bmg{\theta}} = \arg\min_{\bmg{\theta}}
  ~\max_{\bx\in X} ~(\textrm{net}(\bx,\bmg{\theta})-t(\bx))^2
\end{equation}
A severe drawback, however, is that instead of optimizing the distance to the
decision boundary in the input space, the largest deviation in the network
output space is optimized. Unless the network is linear, such as a traditional
perceptron, this will yield a significantly different neural net.

\subsection{Support vector machines}
The key principle behind the support vector machine (SVM), the structural risk
minimization leading to the maximum margin classifier, makes it an ideal
candidate for domain learning. Thanks to the reproducing property of kernels,
in the case of non-overlapping classes, the SVM is a maximum margin hyperplane
in a Hilbert space induced by the specified kernel \cite{Vapnik}. The margin is
determined only by support vectors. These are the boundary objects, \ie the
objects closest to the decision boundary $f(\bx,\bmg{\theta})$
\cite{Cristianini00,Vapnik}. As such, the SVM is independent of class density
models:
\begin{equation}\label{eq:svm}
   f(\bx,\bmg{\theta}) =
   \arg \max_{\bmg{\theta}} ~\min_{\bx \in X} ~y(\bx) f(\bx,\bmg{\theta}).
\end{equation}
Multiple copies of the same object added to the training set do not contribute
to the construction of the SVM, as they do for classifiers based on some
probabilistic model. Moreover, the SVM is also not affected if objects which
are further away from the decision boundary are disregarded or if objects of
the same class are added there. This decision function is, thereby, truly
domain based.

For nonlinear classifiers $f(\bx,\bmg{\theta})$ defined on nonlinear kernels,
the SVM has, however, a similar drawback as the nonlinear neural network. The
distances to the decision boundary are computed in the output Hilbert space
defined by the kernel and not in the input space. A second problem is that the
soft-margin formulation \cite{Cristianini00}, the traditional solution to
overlapping classes is not domain based. The optimization problem for a linear
classifier $f(\bx) = \bw^T \bx + w_0$ is rewritten into:
\begin{equation}\label{eq:svm_overlap}
\begin{array}{ll}
 \min_{\bw} & ||\bw||^2 + \sum_{\bx_i \in X} \xi(\bx_i),\\[0.5mm]
 s.t. & y_i f(\bx_i) \ge 1 - \xi(\bx_i), \\[0.5mm]
      & \xi(\bx_i) \ge 0\\
\end{array}
\end{equation}
in which the term $\sum_{\bx_i \in X} \xi(\bx_i)$ is an upper bound of the
misclassification error on the training set, hence it is responsible for
minimizing \emph{a sum of error contributions}. Adding a copy of an erroneously
assigned object will affect the sum and, thereby, will influence the sought
optimum $\bw$. The result is, thereby, dependent on the distribution of
objects, not just on their domain. For a proper domain based solution,
formulation \eqref{eq:svm} should be solved as it is for the case of
overlapping domains, resulting in the {\em negative margin support vector
machine}. This means that the distance of the furthest away misclassified
object should be minimized. As the signed distance is negative, the negative
margin is obtained. In the probabilistic approach this classifier is unpopular
as it will be sensitive to outliers. As explained in the introduction, in
domain learning, the existence of outliers should be neglected. This implies
that, if they exist, they should be removed before, as they can only be
detected on distribution information.

\section{Evaluation procedure}\label{eval_proc}
In the previous section a number of possible domain based classifiers has been
discussed, inspired by well known probabilistic procedures. This is just an
attempt to illustrate the key points of domain learning approaches. Some of
them are feasible, like the nearest center rule and the maximum error neural
network. Others seem to be almost intractable as the question of determining
multidimensional domains that fit around a given set of points lead to hard
optimization problems. Dropping the assumption that the probability
distribution of the objects is representative for the distribution of the
target objects to be classified is apparently very significant. The consequence
is that the statistical approach has to be replaced by an estimate of the shape
of the class domains.
\begin{figure}[t]
\begin{center}
\includegraphics[height=6.0cm]{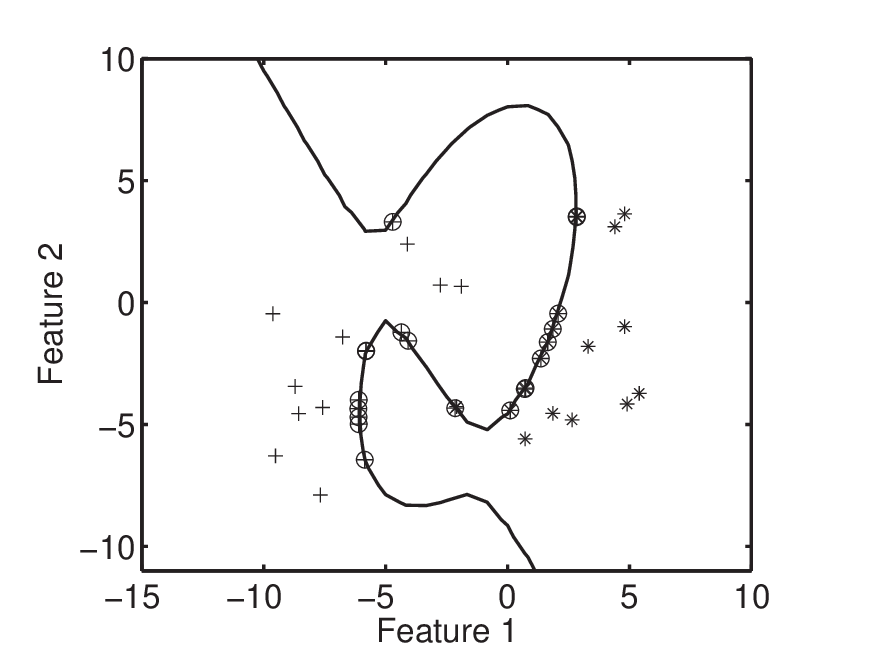}
\vspace*{-1mm} \caption{Example of the projection of a small set of objects on
a nonlinear decision boundary.} \vspace*{-2mm} \label{fig:project_ex}
\end{center}
\end{figure}
The problem of defining consistent classification procedures is not the only
one in domain learning. As it was already noticed, for a proper optimization,
the distance from the objects to the decision boundary or to the domain
boundary should be determined \emph{in the input space}. Here, the original
object representation is defined for the application, so the distances measured
in this space are related in a meaningful way to the differences between
objects. This relation does not hold for the \emph{output space} of nonlinear
decision functions. Still, well-performing classifiers may be obtained. The
question, however, arises how evaluation and a comparison of classifiers that
establish different nonlinearities,\eg a linear classifier, a neural network
and a support vector machine should be done.

The only way various classification functions can be compared is in their
common input space, as their output spaces may differ. In the introduction,
criterion \eqref{eq:criterion} was adopted stating that the performance of a
domain based classifier is determined by the classification of the most
difficult example. It is determined by the distance in the input space from
that object to the decision boundary. For linear classifiers the computation of
this distance is straightforward. For analytical nonlinear classifiers the
computation of this distance is not trivial, but might be defined based on some
optimization procedure over the decision boundary. For arbitrary decision
function, there is no way to derive this distance directly. In order to compare
classifiers of various nature we propose the following heuristic procedure
based on a stochastic approximation of the distance of an object to the
decision boundary:
\begin{enumerate}
\item Let $f$ be a classifier found in the input space $\RR^m$.
Given an independent test set $S$, generate a large set of objects $R \in
\RR^m$ that lie in the neighborhood of the test examples.
\item Label the objects in $S$ and $R$ by the classifier $f$.
\item For each object $\bx_s$ in $S$ find the $k$ nearest objects
$\bx^i_r$ in $R$ that are assigned different labels.
\item Enrich this set $\{\bx^i_r, i\eq 1,2,\ldots,k\}$ by interpolation.
\item Use successive bisection to find the points $\bx^i_c$ that are on the
lines between $\bx_s$ and all $\bx^i_r$ such that they are almost on the
decision boundary induced by $f$.
\item Find the point $\bx_c$ in $\{\bx^i_c\}$ that is nearest to $\bx_s$.
\item Use the distance $d(\bx_s,\bx_c)$ between $\bx_s$ and $\bx_c$ as a measure
for the confidence in the classification of $\bx_s$. If the true label of
$\bx_s$ is known the distance to $\bx_c$ may be given a sign: positive for a
correct label, negative for an incorrect one.
\item Use
\begin{equation}
e_S = min_{\bx_s \in S} d(\bx_s,\bx_c) \label{eq:perf}
\end{equation}
as a performance measure for the evaluated classifier $f$ given the test set
$S$.
\end{enumerate}
This proposed procedure has to be further evaluated. An example of the result
of the projection of a small test set on a given classifier is shown in
fig.\,\ref{fig:project_ex}.
\begin{figure}[t]
\begin{center}
\includegraphics[height=6.0cm]{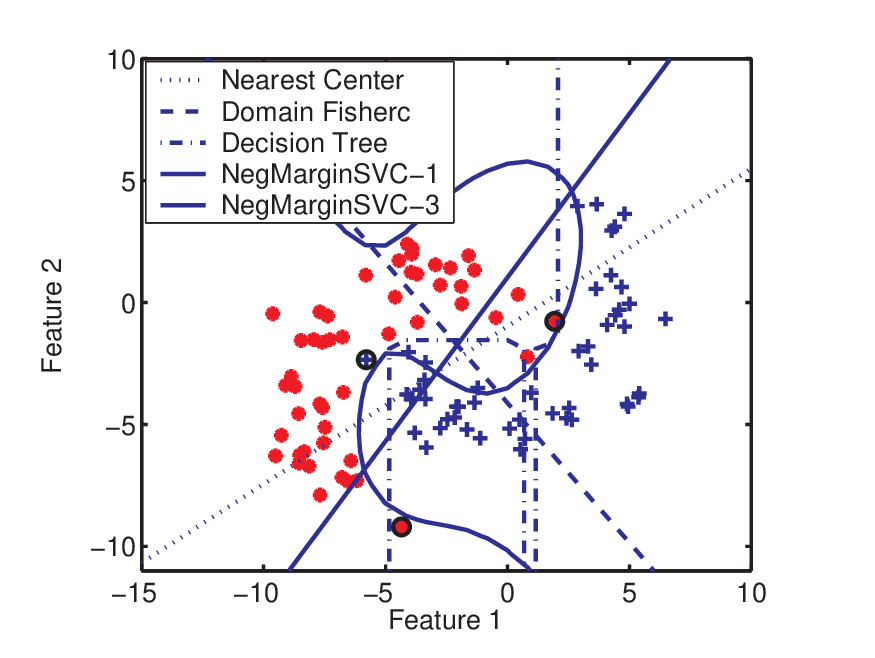}
\vspace*{-1mm} \caption{Five domain based classifiers on artificial data. The
three support objects of the linear Negative Margin SVM are indicated by
circles. } \vspace*{-2mm} \label{fig:art_data_ex}
\end{center}
\end{figure}

\section{Example}
\begin{figure}[t]
\begin{center}
\includegraphics[height=6.0cm]{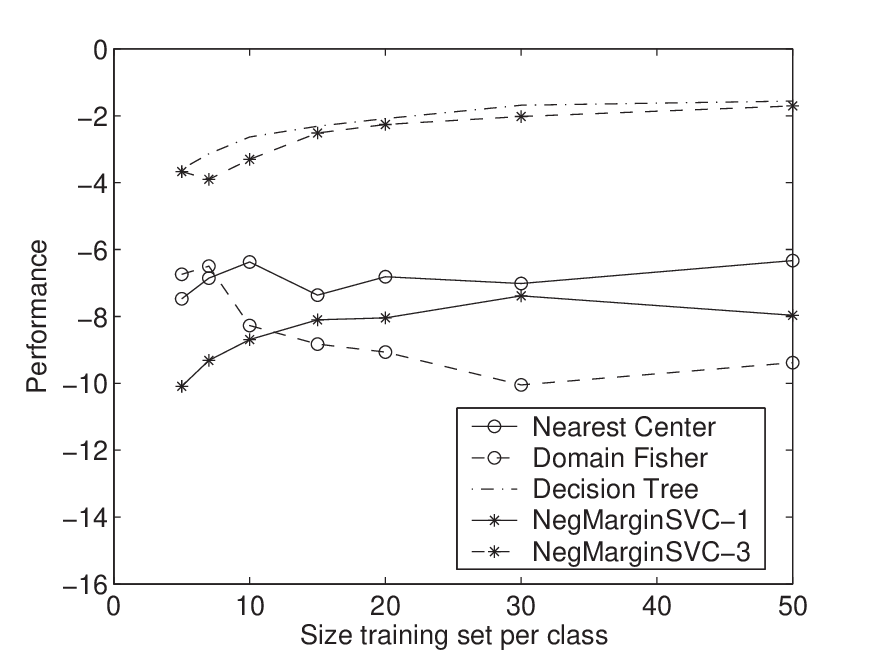}
\vspace*{-1mm} \caption{Learning curves for the five domain based classifiers.
As classes overlap, the performance \eqref{eq:perf} is negative. Higher
performance indicates better results.} \vspace*{-2mm} \label{learning_curves}
\end{center}
\end{figure}

We implemented the following domain based classifiers:
\begin{description}
    \item[Nearest Center,] NCC, based on \eqref{eq:def_center}.
    \item[Domain Fisher] based on
    \eqref{eq:FLDD}, using an heuristic estimate of $G$ by pre-whitening the data
    followed by the NCC to determine the class centers.
    \item[Decision Tree] using the purity criterion.
    \item[Negative Margin SVM] using a linear kernel. As the
    optimization problem is not quadratic, we implemented this
    classifier using boosting \cite{Schapire02}.
    \item[Negative Margin SVM] using a 3rd order polynomial kernel.
\end{description}
Two slightly overlapping artificial banana shaped classes are generated in two
dimensions. Fig.\,\ref{fig:art_data_ex} shows an example for $50$ objects per
class. The decision boundaries for the above mentioned classifiers are also
presented there.

The following experiment is performed using a fixed test set of $200$ examples
per class. Training sets of the cardinalities up to $50$ objects per class are
generated, such that smaller sets are contained in the larger ones. For each
training set the above classifiers are determined and evaluated using the
procedure discussed in section \ref{eval_proc}. This is repeated $10$ times and
the performances is averaged.

Fig.\,\ref{learning_curves} presents the results as a function of the
cardinality of the training set. These are the learning curves of five
classifiers showing an increasing performance as a function of the training
size. As the classes slightly overlap, the performance \eqref{eq:perf} is
negative. This is caused by the fact that the 'worst' classified object in the
test set is erroneously labeled and is, thereby on the wrong side of the
decision boundary.

The curves indicate that our implementation of the Domain Fisher
Discriminant is bad, at least for these data. This might be
understood that it is sensitive for all class boundary points, to
all sides. Enlarging the dataset may yield more disturbances. The
simpler Nearest Center classifier performs much better and is
about similar to the linear SVM. The nonlinear SVM as well as the
Decision Tree yield very good results. Our evaluation procedure
should bad for overlapping training sets classified by the
Decision Tree, as small regions separated out in different classes
disturb the procedure. They are, however, not detected if there
size is really small. The probability that inside such a region a
point is generated (compare the procedure discussed in section
\ref{eval_proc} may be too small.

\section{Conclusions}\label{conclusion}
Traditional ways of learning are inappropriate or inaccurate if
training sets are only representative for the domain, but not for
the distribution of the target objects. In this paper, a number of
domain based classifiers have been discussed. Instead of
minimizing the expected number of classification errors, the
minimum distance to the decision boundary is proposed as a
criterion. This is difficult to compute for arbitrary nonlinear
classifiers. A heuristic procedure based on generating points
close to the decision boundary is proposed for classifier
evaluation.

This paper is restricted to an introduction to domain learning. It
formulates the problem, points towards possible solutions and
gives some examples. A first series of domain based classifiers
has been implemented. Much research has to be done to make the
domain based classification approach ready for applications. As
there is a large need for novel approaches in this area, we
believe that an important new theoretical direction for further
investigation is identified.

\section*{Acknowledgments}
This work is supported by the Dutch Organization for Scientific
Research (NWO).

\bibliographystyle{mlapa}
\bibliography{ref}
\end{document}